\documentclass[runningheads]{llncs}
\usepackage{amsmath}
\usepackage{graphicx}
\usepackage{lipsum}
\usepackage{amssymb}
\DeclareMathOperator*{\argmax}{argmax}
\usepackage{booktabs}
\usepackage{color}
\usepackage{float}
\usepackage{wrapfig}
\usepackage{multirow}
\usepackage{amsfonts}
\usepackage{bbding}

\begin{document}

\title{IRNet: Instance Relation Network for Overlapping Cervical Cell Segmentation}
\titlerunning{IRNet}

\author{Yanning Zhou\inst{1},
Hao Chen (\Envelope)\inst{2},
Jiaqi Xu\inst{1},
Qi Dou\inst{3},
Pheng-Ann Heng\inst{1,4}}


\authorrunning{Y. Zhou et al.}

\institute{$^1$Department of Computer Science and Engineering, The Chinese University of Hong Kong, Hong Kong SAR, China\\
\email{\{ynzhou, hchen\}@cse.cuhk.edu.hk}\\
$^2$Imsight Medical Technology, Co., Ltd. Hong Kong SAR, China\\
$^3$Department of Computing, Imperial College London, London, UK\\
$^4$Guangdong Provincial Key Laboratory of Computer Vision and Virtual Reality Technology, Shenzhen Institutes of Advanced Technology, Chinese Academy of Sciences, Shenzhen, China}

\maketitle              
\begin{abstract}

Cell instance segmentation in Pap smear image remains challenging due to the wide existence of occlusion among translucent cytoplasm in cell clumps.
Conventional methods heavily rely on accurate nuclei detection results and are easily disturbed by miscellaneous objects.
In this paper, we propose a novel Instance Relation Network (IRNet) for robust overlapping cell segmentation by exploring instance relation interaction.
Specifically, we propose the Instance Relation Module to construct the cell association matrix for transferring information among individual cell-instance features.
With the collaboration of different instances, the augmented features gain benefits from contextual information and improve semantic consistency.
Meanwhile, we proposed a sparsity constrained Duplicate Removal Module to eliminate the misalignment between classification and localization accuracy for candidates selection.
The largest cervical Pap smear (CPS) dataset with more than 8000 cell annotations in Pap smear image was constructed for comprehensive evaluation.
Our method outperforms other methods by a large margin, demonstrating the effectiveness of exploring instance relation.

\end{abstract}
\section{Introduction}
Pap smear test is extensively used in cervical cancer screening to assist premalignant and malignant grading~\cite{Pap1942}.
By estimating the shape and morphology structure, e.g., nuclei to cytoplasm ratio, cytologists can give a preliminary diagnosis and facilitate the subsequent treatment.
Given that this work is time-consuming and has large intra-/inter-observer variability, designing automatic cell detection and segmentation methods is a promising way towards accurate, objective and efficient diagnosis.
However, it remains challenging because the multiple layers of cells partially occlude each other in the Pap smear image, while in H\&E image cells do not have multiple layers of translucent overlap.
The widely existence of cell clumps along with the translucent cytoplasm raises obstacles to accurately find the cell boundary.
In addition, apart from the target cervical cells, other miscellaneous instances such as white blood cells, mucus and other artifacts are also scattered in the image, which requires an algorithm robust enough to identify them from targets.

Previously, most of the overlapping cell segmentation methods in Pap smear image utilize the shape and intensity information and can be divided into the following steps: cell clump segmentation, nuclei detection and cytoplasm boundary refinement~\cite{gencctav2012unsupervised,lu2015improved,song2017accurate}.
However, they demand the precise nuclei detection results as the seeds for the further cytoplasm partition and refinement, which is easily disturbed by the mucus, blood and other miscellaneous instances in clinical diagnosis.
Many deep learning based methods have been proposed for gland/nuclei instance segmentation tasks~\cite{chen2017dcan,kumar2017dataset,RAZA2019160}.
Raza et al. proposed Micro-Net for general segmentation task and achieved good results for cell, nuclei and gland segmentation  ~\cite{RAZA2019160}.
But it cannot tackle overlapping instances where one pixel could be assigned to multiple instance IDs. 
On the other hand, the proposal based method can assign multiple labels to a single pixel, which has shown promising results in general object segmentation task.
~\cite{he2017mask} firstly extended the detection method with a segmentation head for instance segmentation, ~\cite{liu2018path} proposed a powerful feature aggregation network backbone.
Akram et al. presented the CSPNet consisting of two sub-nets for proposal generation and segmentation nuclei respectively in microscopic image~\cite{akram2016cell}.
However, in the Pap smear image, directly extracting in-box features from cell clumps for further processing is not informative enough to distinguish the foreground/background cytoplasm fragment.
Meanwhile, the large appearance variance between the single cell and clumps make features semantically inconsistent, which eventually leads to the ambiguous boundary prediction.
Besides, it is easy for greedy Non-Maximum Suppression (NMS) to reject true positive predictions in heavy cluster regions due to the misalignment between classification and localization accuracy.
Motivated by clinical observation that the appearance of each independent cervical cell in Pap smear image has strong similarity, it shows the potential of leveraging relation information which has been shown effectiveness in other tasks ~\cite{NIPS2017_attetnion,hu2018relation,fu2018dual} for better feature representation.
For the first time we introduce relation interaction to instance segmentation task and present the Instance Relation Network (IRNet) for overlapping cervical cell segmentation.
A novel Instance Relation Module (IRM) is introduced which computes the class-specific instance association for feature refinement.
By transferring information among instances using self-attention mechanism, the augmented feature takes merit of contextual information and increase semantic consistency.
We also proposed the Duplicate Removal Module (DRM) with the sparsity constraint to benefit proposal selection by calibrating the misalignment between classification score and localization accuracy.
To the best of our knowledge, the IRNet is the first end-to-end deep learning method for overlapping cell segmentation in Pap smear image.

\section{Method}
\begin{figure}[!h]
\centering
	\includegraphics[width=1\textwidth]{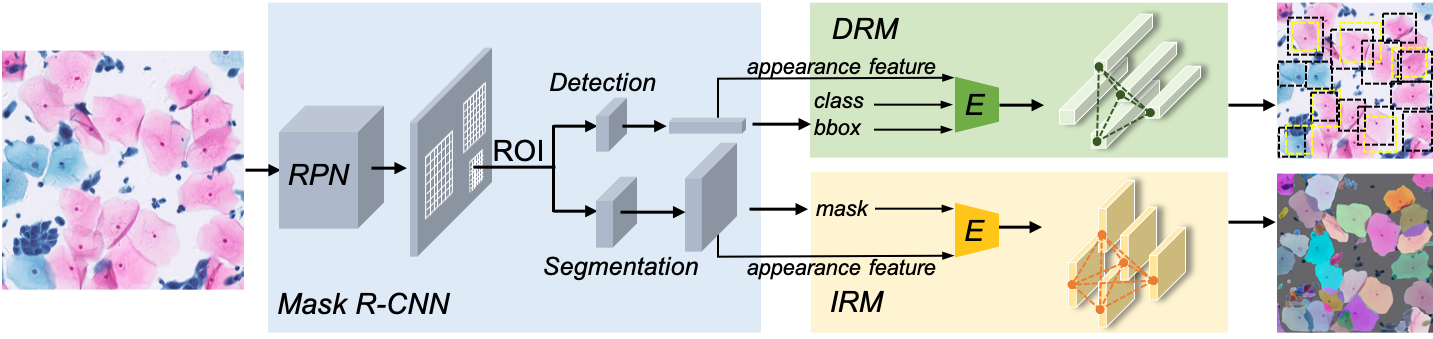}
	\caption{ Overview of the proposed IRNet. }
		\label{framework}
	
\end{figure}

As shown in Fig.~\ref{framework}, the proposed IRNet conforms to the two-stage proposal based instance segmentation paradigm~\cite{he2017mask}.
The input image is firstly fed into the Region Proposal Network (RPN) to generate object candidates.
Then the candidate features are extracted by the RoIAlign layer~\cite{he2017mask} and passed through two branches for detection and segmentation.
To strengthen the network's ability of candidate selection in cell clumps and improve the semantic consistency among in-box features, we leverage the contextual information among different cells by adding Duplicate Removal Module (DRM) and Instance Relation Module (IRM) after detection and segmentation head.

\subsection{Instance Relation Module}

\begin{figure}[h]
	\includegraphics[width=1\textwidth]{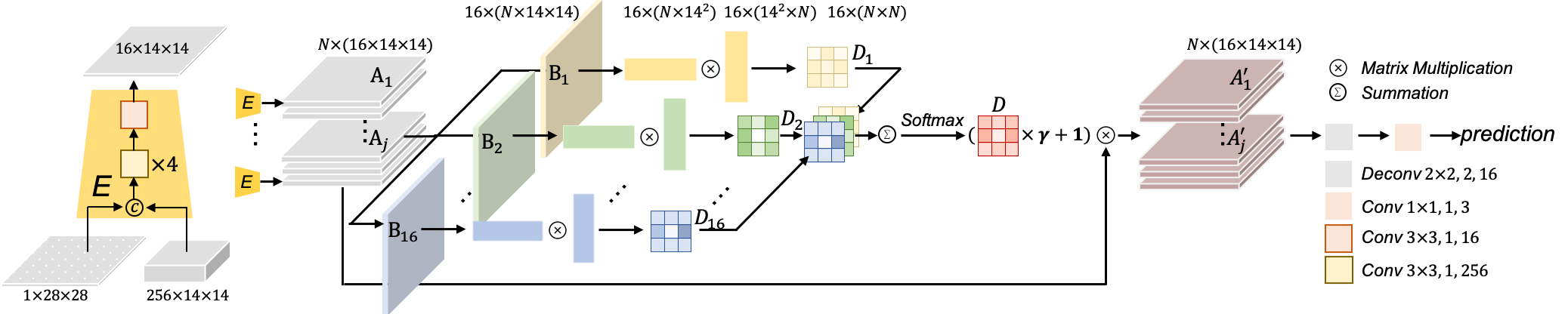}
	\caption{Detail structure of the Instance Relation Module in IRNet.}

		\label{fig:irm}
\end{figure}

Utilizing in-box features to generate each mask independently is susceptible for cell clumps due to the low foreground contrast and the overlapping boundaries, which eventually leads to ambiguous predictions.
Directly enlarging the anchor size to add context won't help a lot in the overlapping region since the surroundings are cells with low contrast. 
Given that nuclei share the strong appearance similarity so as the cells (shape, texture), we hypothesize that utilizing contextual information from other instance as guidance can increase semantic consistency, especially from those well-differentiated cells.
Therefore, we propose the Instance Relation Module (IRM) to exploit the collaborative interaction of instances.
Generally speaking, the IRM takes embedded features from each instance to calculate the instance association matrix, then parses message among features according to their instance relations.

Specifically, the encoder (denotes as \textbf{E} in Fig.~\ref{fig:irm}) takes the combination of the predicted mask and deep features as the input to generate the fused features for each candidate.
Let $n$ denotes the number of instances in the image, the self-attention mechanism~\cite{NIPS2017_attetnion}is used to build the association among $n$ instances.
As can be seen in Fig.~\ref{fig:irm}, the IRM firstly aggregated the fused features in channel-wise to construct 16 features, denoted as $B_{j}$, $j = 1,2 \dots 16$, with the shape of $n \times 14\times 14$.
For each $B_{j}$, it is reshaped to $\mathbb{R}^{c \times hw}$ and multiplied with its transpose matrix to calculate the channel-wise instance associations, $D_{j} = B_{j}B_{j}^{T}$.
The overall instance association matrix is finally obtained by averaging among all the channel-wise association matrices followed by a Softmax layer for normalization, $D = Softmax(avg( D_{1}, D_{2} \dots D_{c}))$.
Therefore, the impact of the $q$-th instance to the $p$-th instance is computed as $ w_{pq} = \frac{exp(d_{pq})}{\sum_{k}exp(d_{pk})} $,
where $d_{pq}$ represents the $p$-th row, $q$-th column entry.
Let $A_{p}$ and $A_{q}$ denotes the $p$-th and the $q$-th instance features, $A_{p}^{'}$ denotes the $p$-th instance features after relation interaction, the message parsing process can be formulated as:
\begin{equation}
    A^{'}_{p} =\gamma\sum _{q=1}^{n}w_{pq}A_{q} + A_{p},
\end{equation}
where $\gamma$ denotes a learnable scalar factor.

By associating with all the instances, the augmented feature takes merit of contextual information from other instance areas to increase semantic consistency.
It is then passed through one $2 \times 2$ deconvolution layer with a stride of two, followed by a convolution layer serving as the classifier to output the predicted masks.
During training, the Binary Cross-Entropy (BCE) loss is calculated on the ground truth class of masks ($\mathcal{L}_{IRM}$).

\subsection{Sparsity Regularized Duplicate Removal Module}

Directly utilizing objectness score for NMS leads to sub-optimal results due to the misalignment between classification and localization accuracy, which is more severe for proposals in cell clumps.
To calibrate the misalignment,~\cite{hu2018relation} proposed the Duplicate Removal Module (DRM) which takes appearance and location features as input and then utilizes transformed features after relation interaction to directly predict the proposal be \textit{correct} or the \textit{duplicate} using BCE loss ($\mathcal{L}_{DRM}$).
The motivation for adding DRM is that the cells and corresponding nuclei have highly correlated spatial distribution.

Based on the observation that cells in Pap smear image gather in several local small clusters instead of one large clump, we propose to add a sparsity constraint on DRM to let the module focus on interaction among the subset of proposals.
Specifically, for each target, it only takes proposals with relation weight ranked in the top $k$ for message parsing, where we set $k=40$ in the experiments.
Meanwhile, instead of directly utilizing predicted probability for duplicate removal, we use the multiplication of classification score and predicted probability for NMS to give a hard constraint of overlapping ratio.
Notice that the DRM also exploits the relation information. The proposed IRM is significantly different to adapt to instance segmentation by utilizing a fully convolutional encoder to combine features and predicted masks which effectively preserves the location information and strengthens the effort of shape information.

\subsection{Overall Loss Function and Optimization}

A multi-task loss is defined as $\mathcal{L} = \mathcal{L}_{cls} + \mathcal{L}_{reg} + \mathcal{L}_{seg} +\alpha\mathcal{L}_{DRM} +\beta\mathcal{L}_{IRM} $, where $\mathcal{L}_{cls}$ and $\mathcal{L}_{reg}$ denote the BCE loss and smooth L1 loss for classification and regression in detection head, and $ \mathcal{L}_{seg}$ denotes pixel-wise BCE loss in segmentation head, which are identical as those defined in~\cite{he2017mask}.
$\mathcal{L}_{DRM}$ denotes the BCE loss for \textit{correct} or \textit{duplicate} classification in DRM, where we define the \textit{correct} as the predicted bounding box with the maximum Intersection over Union to the corresponded grounding truth, while others are \textit{duplicate}.
$L_{IRM}$ is the pixel-wise BCE loss for refined masks after IRM.
$\alpha$ and $\beta$ are hyper-parameters term for balancing loss weights. 

\section{Experiments and Results}
\noindent\textbf{Dataset and evaluation metrics.}
The liquid-based Pap test specimen was collected from 82 patients to build the CPS dataset.
The specimen was imaged in $\times 40$ objective to give the image resolution of around 0.2529 $\mu m$ per pixel.
Then they were cropped into 413 images with the resolution of $1000 \times 1000$.
In all, 4439 cytoplasm and 4789 nuclei were annotated by the cytologist.
To the best of our knowledge, there is no public cervical cell dataset with the annotations on the same order of magnitude to the CPS dataset.
To evaluate the proposed method, we split the dataset in patient-level with the ratio of 7:1:2 into the train, valid and test set.

For quantitative evaluation, Average Jaccard Index (AJI) is used which considers in both pixel and object level~\cite{kumar2017dataset}. AJI$=\frac{\sum_{i=1}^{n}G_{i}\bigcap S_{j}}{\sum_{i=1}^{n}G_{i}\bigcup S_{j}+\sum_{k\in N}S_{k}}$ , where $G_{i}$ is the $i$-th ground truth, $S_{j}$ is the $j$-th prediction, $j = \argmax_{k} \frac{G_{i}\bigcap S_{k}}{G_{i}\bigcup  S_{k}}$.
It measures the ratio of the aggregated intersection and aggregated union for all the predictions and ground truths in the image.
F1-score (F1) is used to measure the detection accuracy for reference~\cite{chen2017dcan}.
\noindent\textbf{Implementation details.}
We implemented the proposed IRNet with PyTorch 1.0.
The network architecture is the same as~\cite{lin2017feature} in the condition of Feature Pyramid Network with 50-layer ResNet (ResNet-50-FPN).
One NVIDIA TITIAN Xp graphic card with CUDA 9.0 and cuDNN 6.0 was used for the computation.
During training, we used SGD with 0.9 momentum as the optimizer.
The initial learning rate was set as 0.0025 with a factor of 2 for the bias, while the weight decay was set 0.0001.
We linear warmed up the learning rate in the first 500 iterations with a warm-up factor of $\frac{1}{3}$.

\noindent\textbf{Effectiveness of the proposed IRNet.}
Firstly, we conducted experiments to compare different algorithms for overlapping cell segmentation.
(1).\textit{JOMLS}~\cite{lu2015improved}: an improved joint optimization of multiple level set functions for the segmentation of cytoplasm and nuclei from clumps of overlapping cervical cells. 
(2).\textit{CSPNet}~\cite{akram2016cell}: a cell segmentation proposal network with two CNNs for cell proposal prediction and cell mask segmentation respectively.
To give a fair comparison, we reproduce the ResNet-50-FPN instead of the original 6-layer CNN for candidates prediction in the first stage.
(3). \textit{Mask R-CNN}~\cite{he2017mask}: IRNet without Duplicate Removal Module and Instance Relation Module, which can be considered as a standard mask-rcnn structure with ResNet-50-FPN as the backbone.
(4).\textit{IRNet w/o IRM}: IRNet without Instance Relation Module.
(5). \textit{IRNet w/o DRM}: IRNet without Duplicate Removal Module.
(6). \textit{IRNet}: The proposed IRNet.

\begin{table}[t]
    \centering
        \caption{Quantitative comparison against other methods on the test set.}
    \setlength{\belowcaptionskip}{-10pt}
    \begin{tabular}{p{3cm}|p{1.4cm}p{1.4cm}p{1.4cm}p{1.4cm}}
    \toprule
    \hfil\multirow{2}{*}{ Method} & \multicolumn{2}{c}{AJI} & \multicolumn{2}{c}{F1} \\
    \cline{2-5}

          & \hfil Cyto &\hfil Nuclei & \hfil Cyto& \hfil Nuclei \\
        \hline
         \hfil JOMLS~\cite{lu2015improved}& \hfil0.1974 &\hfil0.3167  & \hfil 0.3794 & \hfil0.3618 \\
        \hfil CSPNet~\cite{akram2016cell}&\hfil0.4607 & \hfil 0.3891 & \hfil0.5307& \hfil 0.5942  \\
\hfil Mask R-CNN~\cite{he2017mask}  &\hfil 0.6845 &\hfil0.5169  &\hfil 0.6664 & \hfil 0.7192   \\
       \hfil  IRNet w/o IRM &\hfil0.6887  &\hfil 0.5342 & \hfil0.7266& \hfil 0.7424 \\

                \hfil IRNet w/o DRM &  \hfil0.6995 &  \hfil0.5471  & \hfil0.7010  & \hfil0.7501 \\


          \hfil \textbf{IRNet}&\hfil \textbf{0.7185} &\hfil \textbf{0.5496}  & \hfil\textbf{0.7497}& \hfil\textbf{0.7554}\\
         \toprule
    \end{tabular}

    \label{tab:tb1}
\end{table}

As can be seen in Table~\ref{tab:tb1}, all the deep learning based methods achieve better performance compared with the level-set based method~\cite{lu2015improved}.
The reason is that our dataset has more complicated background information including white blood cells and other miscellaneous instances compared with that used in~\cite{lu2015improved}, which requires the algorithm to be more robust for the noise.
Apart from CSPNet~\cite{akram2016cell} that uses a separate CNN to extract multi-level features in ROI-pooling, our baseline model extracts features from the shared FPN backbone in specific resolution according to the box size, which effectively improves cytoplasm AJI from 0.4607 to 0.6845 and nuclei AJI from 0.3891 to 0.5169.
In addition, adding the DRM (\textit{IRNet w/o IRM}) gives a striking $9.03\% $ and $3.23\%$ improvement of F1 score for cytoplasm and nuclei, which proves that the duplicated boxes can be effectively suppressed by parsing message among boxes, especially for the cytoplasm in cell clumps.
Compared with baseline model (\textit{Mask R-CNN}), adding IRM after segmentation branch (\textit{IRNet w/o DRM}) gains $2.19\%$ and $5.84\%$ AJI improvements for cytoplasm and nuclei, demonstrates that by leveraging instance association for transferring information, the augmented features are more consistent and discriminative for semantic representation.
By combining DRM and IRM in our framework, the proposed IRNet (\textit{Ours}) outperforms other methods by a large margin, with $4.97\%$ and $6.33\%$ improvements of AJI as well as $12.5\%$ and $5.03\%$ improvements of F1 score for cytoplasm and nuclei comparing with the baseline model (\textit{Mask R-CNN}).

\noindent\textbf{Ablation study for the Instance Relation Module.}
We then conduct the ablation study for the design of the proposed IRM.
(1). \textit{DF} (Deep Feature): Using deep features before the deconvolution layer in segmentation branch with the shape of $256 \times 14 \times 14$ for further process.
(2). \textit{MSK} (Mask): Using predicted mask from segmentation branch for further process.
We keep the masks from the predicted class in detection branch only and remove the other class.
(3). \textit{RL} (Relation Learning): Conducting channel-wise instance relation learning.
Results are shown in Table~\ref{tab:ablation}.

We first add the same number of convolution layers as that in IRM encoder and remove the relation interaction part (\textit{DF + MSK}).
Adding more parameters to build deeper layers do improve the results, but all the methods with relation learning have better performance.
Compared with \textit{IRNet w/o IRM}, directly utilizing deep features for relation interaction (\textit{DF+RL}) yields the results of 0.7097 and 0.5367 in AJI, which brings $3.05\%$ and $0.47\%$ improvement for cytoplasm and nuclei.
Meanwhile, directly using predicted masks (\textit{MSK+RL}) outperforms \textit{IRNet w/o IRM} by $2.67\%$ AJI for cytoplasm.
The nuclei class does not improve in \textit{MSK+RL} because the shape of masks are almost the same so that it does not contain enough information for message parsing.
Furthermore, when we adopt deep features with selected masks simultaneously, it significantly improves the performance over the (\textit{DF+RL}) by $4.33\%$ and $2.88\%$ for cytoplasm and nuclei, which shows the effectiveness of the IRM design.

\begin{table}[t]

    \centering
\footnotesize
    \begin{tabular}{p{1.1cm}p{1.1cm}p{1.1cm}|p{1.4cm}p{1.4cm}p{1.4cm}p{1.4cm}}
    \toprule
      \hfil \multirow{2}{*}{\textbf{ DF}}  &\hfil\multirow{2}{*}{\textbf{ MSK}} &\hfil\multirow{2}{*}{\textbf{ RL}} &\multicolumn{2}{c}{\textbf{ AJI}}&\multicolumn{2}{c}{\textbf{ F1}}  \\
\cline{4-7}
    &&  &\hfil Cyto& \hfil Nuclei&\hfil Cyto& \hfil Nuclei  \\
\cline{1-7}
               &  &&       \hfil0.6887  &\hfil  0.5342  & \hfil0.7266& \hfil 0.7424   \\

              \hfil \checkmark  & \hfil \checkmark &&\hfil0.7042& \hfil0.5283  &\hfil0.7355 & \hfil0.7306   \\

          \hfil \checkmark   & &   \hfil \checkmark  &\hfil0.7097& \hfil0.5367  &\hfil0.7350 & \hfil 0.7442  \\

    & \hfil \checkmark  & \hfil \checkmark&  \hfil0.7071 & \hfil0.5299 & \hfil0.7324 &  \hfil0.7366\\

       \hfil \checkmark  & \hfil \checkmark & \hfil \checkmark&\hfil \textbf{0.7185} &\hfil \textbf{0.5496}    & \hfil\textbf{0.7497}& \hfil\textbf{0.7554}\\

\toprule
    \end{tabular}
    \caption{Ablation study for the Instance Relation Module.}
    \label{tab:ablation}
   
\end{table}

\noindent\textbf{Qualitative comparison.}
\begin{figure}[t]
\centering
	\includegraphics[width=1\textwidth]{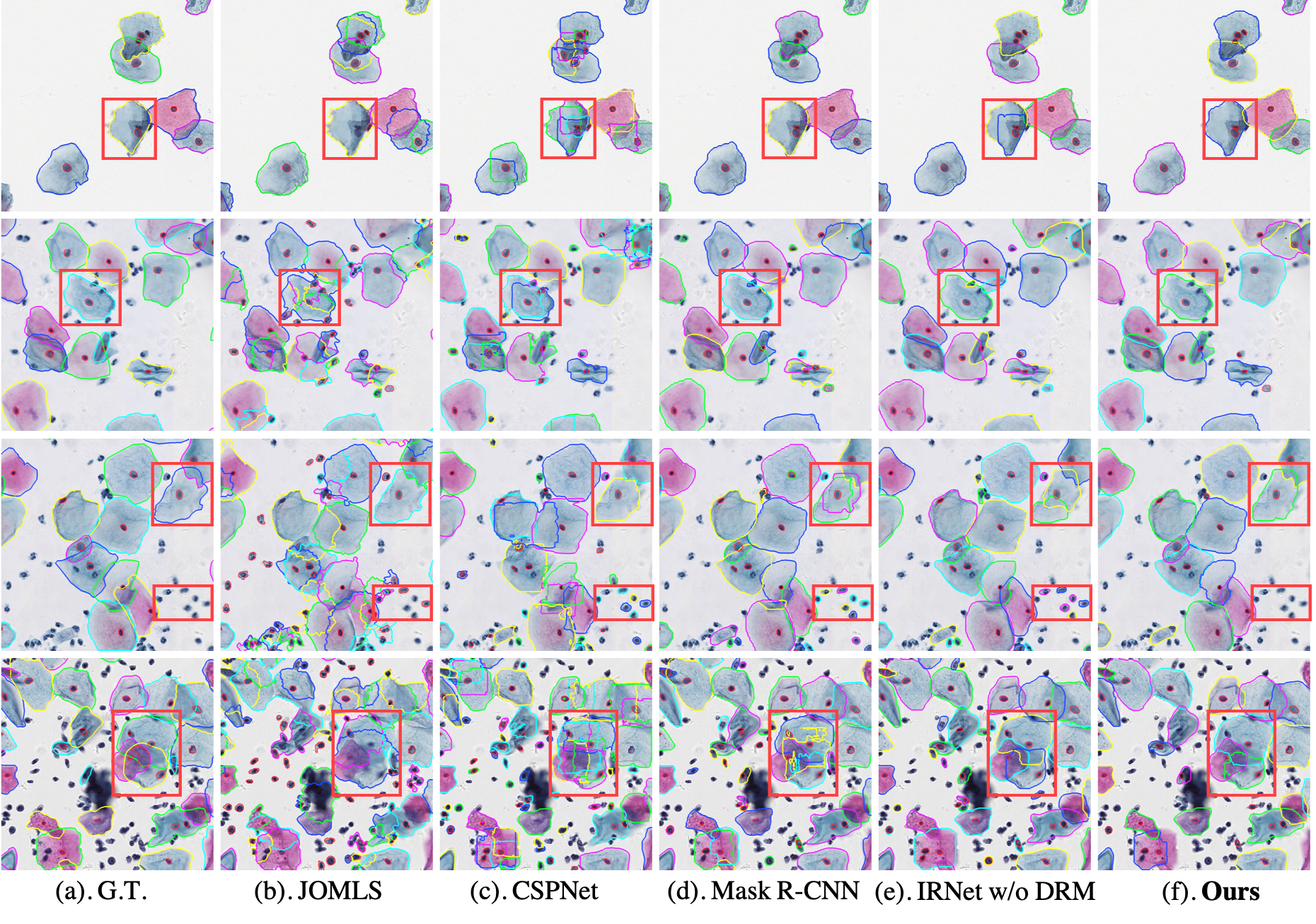}
	\caption{Qualitative results of Overlapping cervical cell segmentation in Pap smear image on the test set (each closed curve denotes an individual instance). Rectangles show the main differences among different methods.}
		\label{visualization}

\end{figure}
Fig. \ref{visualization} shows representative samples in the test set with challenging cases such as the heavily occlusion of cytoplasm and the scatted white blood cells.
Conventional method (\textit{JOMLS}) fails on identifying the miscellaneous instance and the nucleus (see the third row and fourth row).
Compared with \textit{Mask R-CNN}, adding IRM mitigates the ambiguous cytoplasm boundary prediction in cell clumps (see (a) and (e)), which is important for further accurate cell classification by morphology analysis.
Moreover, combining DRM further suppresses the duplicated predictions successfully (see (e) and (f)).

\section{Conclusion}
Accurately segmenting the cytoplasm and nucleus instance in Pap smear image is pivotal for further cell morphology analysis and cervical cancer grading.
In this paper, we proposed a novel IRNet which leverages the instance relation information for better semantic consistent feature representation.
By aggregating the features from other instances,they tend to generate masks with more consistent boundary shape.
Quantitative and qualitative results demonstrate the effectiveness of our method.
Notice the proposed IRM is inherently general and can be complementary for various proposal-based instance segmentation methods.

\section{Acknowledgement}
This work was supported by 973 Program (Project No. 2015CB351706) grants from the National Natural Science Foundation of China with Project No. U1613219, Research Grants Council of Hong Kong Special Administrative Region under Project No. CUHK14225616, Hong Kong Innovation and Technology Fund under Project No. ITS/041/16, and Shenzhen Science and Technology Program (No.JCYJ20180507182410327).

    \bibliographystyle{splncs04}
    \bibliography{samplepaper}
\end{document}